\def\year{2021}\relax
\documentclass[letterpaper]{article} 
\usepackage{aaai21}  
\usepackage{times}  
\usepackage{helvet} 
\usepackage{courier}  
\usepackage{graphicx} 
\usepackage{natbib}  
\usepackage{caption} 
\nocopyright
\usepackage{times}
\usepackage{latexsym}
\usepackage{color}

\usepackage{booktabs}
\usepackage{multirow}

\usepackage{microtype}
\usepackage{amsmath}\usepackage{amsfonts,amssymb}

\title{DialogBERT: Discourse-Aware Response Generation via Learning to Recover and Rank Utterances}

\author{
   Xiaodong Gu\textsuperscript{\rm 1,2}, Kang Min Yoo\textsuperscript{\rm 2} and Jung-Woo Ha\textsuperscript{\rm 2}\\
  }
  
\affiliations{
    \textsuperscript{\rm 1}School of Software, Shanghai Jiao Tong University\\  \textsuperscript{\rm 2}NAVER AI Lab\\
    xiaodong.gu@sjtu.edu.cn, \{kangmin.yoo, jungwoo.ha\}@navercorp.com\\
}

\date{}

\begin{document}
\maketitle
\begin{abstract}
Recent advances in pre-trained language models have significantly improved neural response generation. 
However, existing methods usually view the dialogue context as a linear sequence of tokens and learn to generate the next word through token-level self-attention. 
Such token-level encoding hinders the exploration of discourse-level coherence among utterances. 
This paper presents DialogBERT, a novel conversational response generation model that enhances previous PLM-based dialogue models. DialogBERT employs a hierarchical Transformer architecture. 
To efficiently capture the discourse-level coherence among utterances, we propose two training objectives, including masked utterance regression and distributed utterance order ranking in analogy to the original BERT training. 
Experiments on three multi-turn conversation datasets show that our approach remarkably outperforms the baselines, such as BART and DialoGPT, in terms of quantitative evaluation. 
The human evaluation suggests that DialogBERT generates more coherent, informative, and human-like responses than the baselines with significant margins.
\end{abstract}

\section{Introduction}
Multi-turn open-domain dialogue modeling is an active research topic in the field of natural language processing. 
However, generating a coherent and informative response for a given dialogue context remains a challenge.
A critical challenge is the learning of rich and robust context representations of dialogue utterances~\citep{ortega2017neural,pragst2018vector}, namely the challenge of encoding a dialogue context into a vector that adequately captures the semantics (e.g., topic, intention).

Large-scale pre-training language models using Transformer-based architectures have recently achieved remarkable successes in a variety of NLP tasks~\citep{devlin2018bert,sun2019ernie, zhang2019hibert,zheng2020pre}.
As such, there are increasingly work that aims to use pre-training language models for conversation modeling~\citep{mehri2019pretraining,zhang2019dialogpt,rothe2019bert2bert}.
For example, DialoGPT~\citep{zhang2019dialogpt} extends the GPT-2~\citep{radford2019language} to generate conversation responses on large-scale dialogue corpus.
Meena~\citep{adiwardana2020meena} trains a sequence-to-sequence model~\citep{cho2014seq2seq} with the Evolved Transformer~\citep{so2019evol} on large-scale multi-turn conversations. 
Blender, developed by Facebook, provides recipes for building open-domain chatbots that perform well in human evaluations~\citep{roller2020recipes}.

However, existing pre-training conversation models usually view the dialogue context as a linear sequence of tokens and learns to generate the next word through token-level self-attention. 
One issue of this approach is that the high-level relationships between utterances are harder to capture using word-level semantics.
For example, the discourse-level relationship between the utterances \emph{``coffee please''} and \emph{``here you are''} (Figure~\ref{fig:arch}) is apparent, but word-level comparisons, such as $<$\emph{coffee}, \emph{you}$>$ and $<$\emph{please}, \emph{are}$>$, obscures the high-level relationship.
Furthermore, this full pairwise attention is inefficient since it requires each word in the context and the decoder to interact with all other words regardless of their distances and semantic units.

To alleviate the issues above, we present DialogBERT, a novel conversational response generation model. 
DialogBERT employs a hierarchical Transformer architecture to represent the dialogue context. 
It first encodes dialogue utterances through a Transformer encoder and then encodes the resulting utterance vectors using a discourse-level Transformer to obtain a representation of the entire dialogue context. 
To efficiently capture discourse-level coherence among utterances, we propose two training objectives in analogy to the original BERT training: 1) \textbf{masked context regression}, which masks a randomly-selected utterance and predicts the encoding vector for the masked utterance directly; and 2) \textbf{distributed utterance order ranking}, which 
organizes randomly shuffled utterances of a conversation into a coherent dialogue context 
through a \emph{Learning-to-Rank}~\citep{cao2007listnet} neural network.

We evaluate DialogBERT on popular multi-turn conversation datasets, namely Weibo, MultiWOZ and DailyDialog. 
Results show that DialogBERT outperforms baselines in terms of perplexity, BLEU, and NIST.
Human evaluation supports the superiority of our approach in capturing discourse-level semantics and generating more plausible dialogue responses.


\section{Related Work}
This work is closely related to (1) pre-trained language models, (2) pre-trained models for conversations, and (3) adopting auxiliary multi-task objectives for improving pre-trained language models.

\noindent\textbf{Pre-trained Language Models}. 
The current paradigm has gradually evolved from word vectors~\citep{pennington2014glove} and contextualized word embedding models~\citep{peters2018deep}.
Recent works have explored various architecture choices and training objectives for large-scale pre-trained language models~\citep{devlin2018bert, radford2019language, yang2019xlnet} based on Transformers~\citep{vaswani2017attention}. 
A recent work proposed using the denoising autoencoder framework with composite corruption schemes~\citep{lewis2019bart}.

\noindent\textbf{Pre-trained Models for Dialogue Generation}. 
Recent advances in pre-trained language models have spurred success in dialogue response generation.
Specifically, \citet{budzianowski2019hello} explored the use of pre-trained language Transformers for task-oriented dialogues.
\citet{wolf2019transfertransfo} also proposed adopting auxiliary unsupervised objectives for pre-training dialogue language models.
Extracting language representations from pre-trained transformers for dialogue tasks has been explored in \citet{henderson2019convert}.

Another important line of work pertains to designing specific Transformer-based architectures that captures dialogue structures and directly pre-training these architectures on dialogue corpora.
\citet{mehri2019pretraining} proposed a Transformer-based hierarchical model and various unsupervised objectives for pre-training contextual semantics of dialogue utterances.
DialogBERT differs from the methods proposed by \citet{mehri2019pretraining} in both the architecture and objectives. 
DialogBERT contains a context encoder that models the discourse coherence, while \citet{mehri2019pretraining} proposed optimizing the utterance encoder directly. 

Lastly, an emerging trend in dialogue generation explores the feasibility of directly pre-training Transformer-based language modeling architectures on large-scale dialogue corpora.
Recent works, such as DialoGPT~\cite{zhang2019dialogpt}, Meena~\cite{adiwardana2020meena}, and Blender~\cite{roller2020recipes}, demonstrate strong generation performances attainable from training Transformer-based language generators on open-domain discourses. 

\noindent\textbf{Multi-task Learning for Pre-training}.
Our work is also profoundly related to auxiliary multi-task learning, which augments the pre-training of language models.
The common theme is to guide the language modeling Transformers with complementing objectives. 
One way is to augment language modeling with annotation rationales~\citep{melamud2019combining}.
ERNIE2~\citep{sun2019ernie} is a continual multi-task learning framework for language understanding that combines unsupervised and supervised objectives.
HIBERT~\citep{zhang2019hibert} has been proposed to model documents using a hierarchical BERT architecture trained with the masked sentence decoding scheme, where the goal is to predict the entire erased sentence.
Our work differs from HIBERT in that (1) we directly match the context sensitive sentence representations with the real utterance encoding and (2) we propose a novel objective that trains the model to predict utterance orders.
ELECTRA~\citep{clark2020electra} explores combining discriminative and generative objectives for learning language modeling.

\section{Approach}
\label{sec:approch}
Let $\mathcal{D}$ = $(u_1, u_2,\ldots, u_T)$ denote a dialogue, where $\mathcal{C}$ = ($u_1, u_2,\ldots, u_{T-1}$) is the dialogue context (history) and $u_T$ is the response. Each $u_i$ = $(w_1^i, w_2^i,\ldots, w_{|u_i|}^i)$ in $\mathcal{C}$ is an utterance and $w_j^i$ is the $j$-th word in $u_i$. Given the context~$\mathcal{C}$, our goal is to generate the next utterance (response)~$u_T$ that is coherent to~$\mathcal{C}$. Consequently, there are two objectives we want to achieve: 1) learning to represent the dialogue context~$\mathcal{C}$ and 2) learning the conditional probability of generating $u_T$ given $\mathcal{C}$.

\subsection{Hierarchical Transformer Encoder}\label{ss:hbert} 
 

	\begin{figure} [!tb]
		\centering 
		\includegraphics[width=3in]{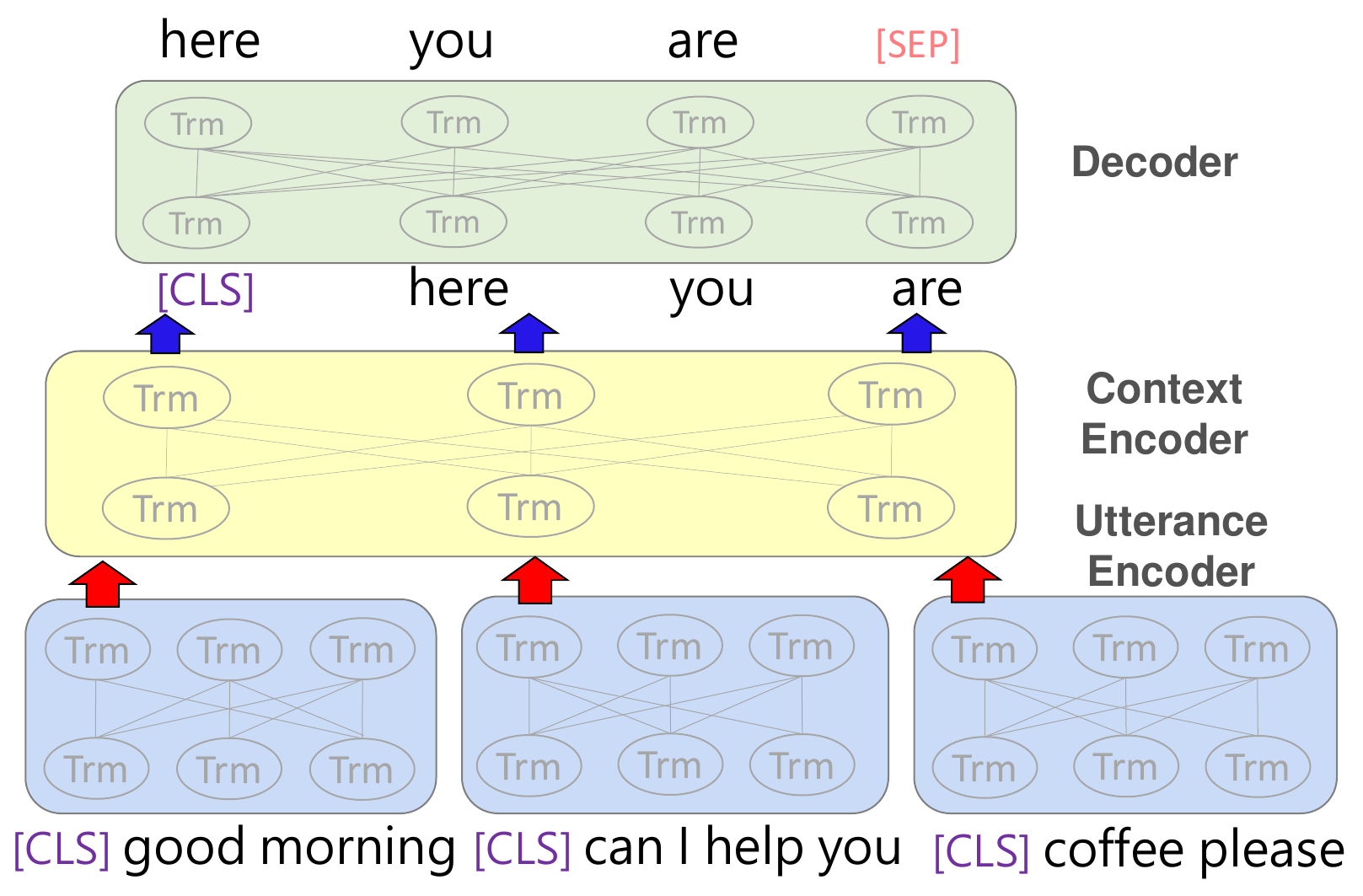} 
		\caption{The hierarchical Transformer encoder-decoder architecture. We omit the [\texttt{SEP}] token in the end of each utterance for simplicity.}
		\label{fig:arch}
	\end{figure}
	
To obtain a better representation of $\mathcal{C}$, we employ a hierarchical transformer encoder architecture. As shown in Figure~\ref{fig:arch}, two transformer encoders are hierarchically nested: an utterance encoder~$f_\theta(\cdot)$ to transform each utterance in $\mathcal{C}$ to a vector and a context encoder~$g_\phi(\cdot)$ to learn utterance representations given their surrounding utterances in the context. Both encoders are based on the Transformer encoder described in \citep{vaswani2017attention}. 

For each $u_i$ = $(w^i_1, w^i_2,\ldots, w^i_{|u_i|})$, inspired by the language modeling pre-training strategy of BERT~\citep{devlin2018bert}, we add [\texttt{CLS}] and [\texttt{SEP}] tokens at the first and the last positions, respectively. Hence, $w_1^i$=[\texttt{CLS}] and $w_{|u_i|}^i$=[\texttt{SEP}]. 
Then, an embedding layer maps $u_i$ onto a continuous space:
\begin{equation}
    \mathbf{e}_i=(\mathbf{w}^i_1+\mathbf{p}_1, \mathbf{w}^i_2+\mathbf{p}_2,\ldots, \mathbf{w}^i_{|u_i|}+\mathbf{p}_{|u_i|})
\end{equation} 
where $\mathbf{w}^i_j$ and $\mathbf{p}_j$ are the word and positional embeddings of $w^i_j$, respectively. 
Then, the utterance encoder~$f_\theta(\cdot)$ transforms $\mathbf{e}_i$ into a list of hidden representations~($\mathbf{u}^i_1$, $\mathbf{u}^i_2$,\ldots, $\mathbf{u}^i_{|u_i|}$): 
  \begin{equation}
      \mathbf{u}^i_1,\ldots,\mathbf{u}^i_{|u_i|} = f_\theta(\mathbf{e}_i).
  \end{equation}
We take the first hidden representation $\mathbf{u}^i_1$ (i.e., the representation at the [\texttt{CLS}] token) as the representation of utterance $u_i$. 
Similar to the representation of each word in $u_i$, we also take the positions of utterances into account. The final representation of $u_i$ is $\mathbf{u}_i$ = $\mathbf{u}^i_1 + \mathbf{p}_i$.

In analogy to the utterance encoder, the context encoder~$g_\phi(\cdot)$ is another Transformer encoder that is applied on the utterance level. As shown in Figure~\ref{fig:arch}, after encoding the utterances, the context encoder transforms the sequence of utterance representations $(\mathbf{u}_1, \mathbf{u}_2,\ldots, \mathbf{u}_{|\mathcal{C}|})$ into context sensitive utterance representations $(\mathbf{h}_1, \mathbf{h}_2,\ldots, \mathbf{h}_{|\mathcal{C}|})$:
     \begin{equation}
      \mathbf{H} = \mathbf{h}_1,\ldots,\mathbf{h}_{|\mathcal{C}|} = g_\phi(\mathbf{u}_1,\ldots,\mathbf{u}_{|\mathcal{C}|})
      \label{eq:uttrep}
  \end{equation}
Equation~\ref{eq:uttrep} is the final product of encoding the dialogue context with a hierarchical bidirectional transformer encoder.

The hierarchical Transformer architecture above is reminiscent of HIBERT in document generation~\cite{zhang2019hibert}. Although both methods encode texts in a hierarchical manner, our objectives and training methods are significantly different. HIBERT is designed for document summary and is only trained through word-by-word decoding of masked sentences in documents. By contrast, DialogBERT has two new loss terms specifically designed for dialog coherence modeling, i.e., the masked utterance regression which is a discourse level regression of masked utterances and the distributed utterance order ranking that reorders randomly-shuffled utterances.

In the next section we will introduce these unsupervised objectives for training DialogBERT.

\subsection{Training Objectives}
  \label{ss:approach:objective}
  In order to capture the discourse-level coherence in dialog contexts, we propose two novel  objectives inspired by BERT in addition to the conventional objective of response generation, i.e., maximizing the log probabilities of decoded words. 
  Following three subsections describe the objectives in turn.
  
\subsubsection{Next Utterance Generation (NUG)}
As the primary goal of response generation, our first training objective is to generate the  next utterance~(response) given the dialog context.
As is shown in Figure~\ref{fig:arch}, we first apply the hierarchical encoder to the context~$\mathcal{C}$ and obtain its context sensitive utterance representations $(\mathbf{h}_1, \mathbf{h}_2,..., \mathbf{h}_{|\mathcal{C}|})$. Then, we generate the next utterance $u_T$ = $(w^T_1,\ldots, w^T_N)$ using a Transformer decoder~$p_\psi(\cdot)$ ($w^T_1$ is also a [\texttt{CLS}] token to be consistent to the utterance encoder). The decoder predicts each word $w^T_j$ conditioned on $w^T_1,\ldots, w^T_{j-1}$ and $\mathbf{h}_1,\ldots,\mathbf{h}_{|\mathcal{C}|}$ by estimating the following probability distribution:
  \begin{equation}
    \label{eq:decoder}
    \begin{split}
      & p(u_{T}|\mathcal{C},\theta,\phi,\psi) = \Pi_{j=1}^N p_\psi(w^T_j|w^T_{<j},\mathbf{H}),
    \end{split}
  \end{equation}
where $N$ represents the maximum sequence length for decoding, while $\theta$, $\phi$, and $\psi$ denote the model parameters of the utterance encoder, the context encoder, and the decoder, respectively.

Finally, the NUG task aims to minimize the cross entropy loss in the decoder:
\begin{equation}
      \mathcal{L}_\mathrm{dec}(\theta,\phi,\psi|w^T_1, \ldots, w^T_N, \mathcal{C}) 
      = - \sum_{i=1}^N \mathrm{log}~p_\psi(w^T_i| w^T_{<i}, \mathbf{H})
  \end{equation}
 where $N$ denotes the maximum sequence length of the generated response.

\subsubsection{Masked Utterance Regression (MUR)}
\begin{figure} [!tb]
		\centering 
		\includegraphics[width=3in]{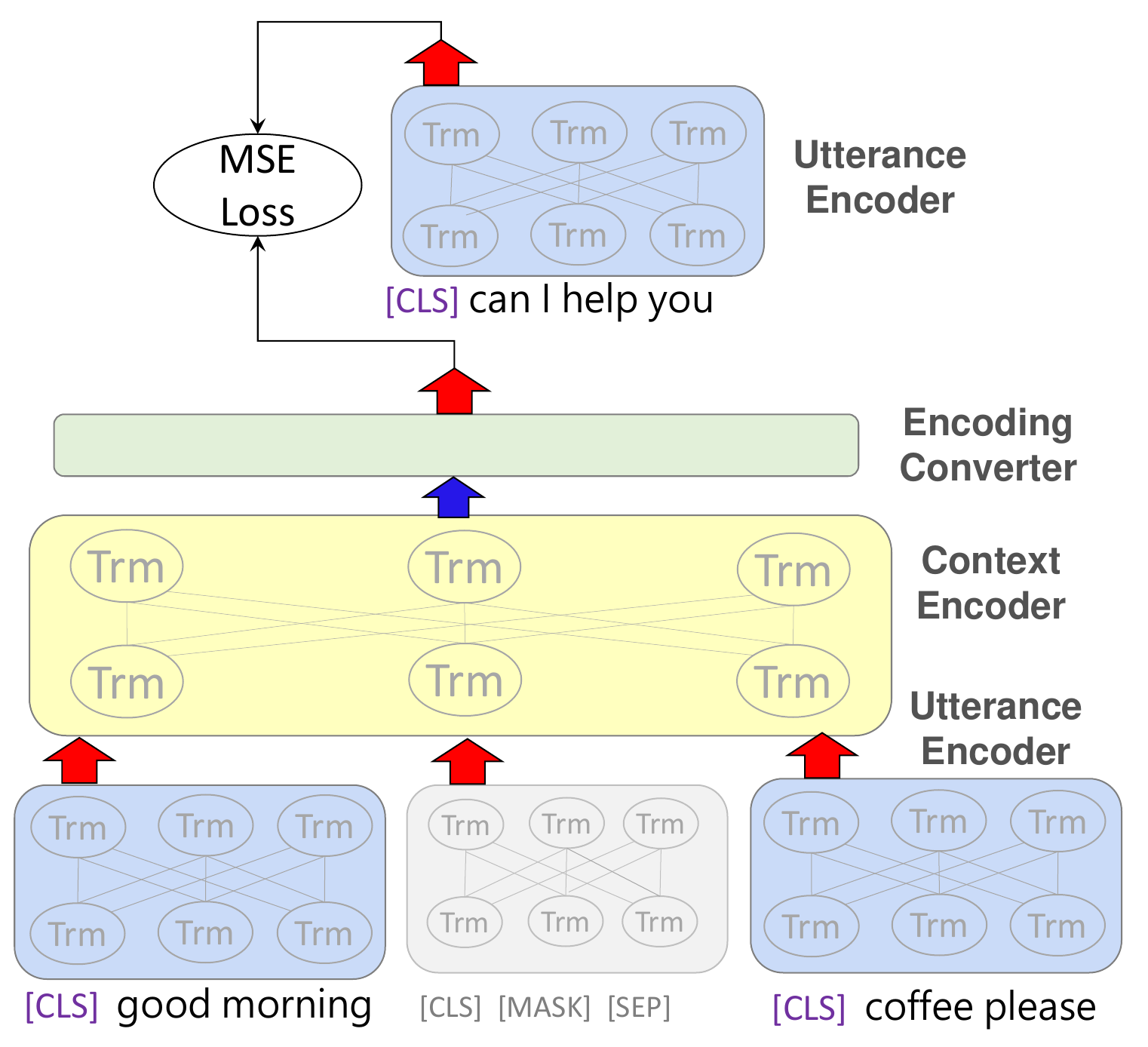} 
		\caption{Illustration of masked utterance regression. The utterance ``\emph{can I help you}'' is masked. We estimate its encoding using an encoding converter given its context-sensitive representation and compare the estimated encoding with the real one produced by the utterance encoder.}
		\label{fig:mur}
	\end{figure}

In analogy to the masked LM~(MLM) in BERT, we design the masked utterance regression as an auxiliary task for enhancing context representation learning (Figure~\ref{fig:mur}). 
Given a dialogue context $\mathcal{C}$ = $(u_1, u_2,\ldots, u_{T-1})$, we randomly select one utterance in $\mathcal{C}$ and the selected utterance is 1) 80\% of time, replaced with a mask utterance~[\texttt{CLS}, \texttt{MASK}, \texttt{SEP}], or 2) 10\% of time unchanged in order to simulate the input context during test time (with no masked utterance), or 3) 10\% of time replaced with a random utterance from the training set. 
We then try to reconstruct the vector of the masked utterance.

After obtaining the masked context $\tilde{\mathcal{C}}$ = $(\tilde{u}_1, \tilde{u}_2,\ldots, \tilde{u}_{|\mathcal{C}|})$, we predict the original utterance vectors from $\tilde{\mathcal{C}}$. We first apply the hierarchical encoder to $\tilde{\mathcal{C}}$ and obtain its context sensitive utterance representations $(\tilde{\mathbf h}_1, \tilde{\mathbf h}_2,\ldots, \tilde{\mathbf h}_{|\mathcal{C}|})$. Then, we transform these representations back to the original utterance vectors using a fully connected neural network:
\begin{equation}
    \hat{\mathbf u}_i=\mathbf{W}\tilde{\mathbf h}_i+ \mathbf{b}
\end{equation}
where $\hat{\mathbf u}_i$ denotes the predicted original utterance vector; $\mathbf{W}$ and $\mathbf{b}$ are trainable parameters. 

Finally, this objective aims to minimize the mean squared error~(MSE) between the estimated representations of masked utterances and their original vectors:
\begin{equation}
  \begin{split}
    & \mathcal{L}_\mathrm{mur}(\theta, \phi, \mathbf{W},\mathbf{b}| \tilde{\mathbf u}_1,\ldots,\tilde{\mathbf u}_{|\mathcal{C}|},\mathcal{C}, \tilde{\mathcal C}, ) \\
    & ~~~~~= \frac{1}{|\tilde{\mathcal C}\setminus \mathcal{C}|}\sum_{u_i\in\tilde{\mathcal C}\setminus\mathcal{C}}|| \hat{\mathbf u}_i -\mathbf{u}_i||_2^2
  \end{split}
\end{equation}
where $\tilde{\mathcal C}\setminus\mathcal{C}$ denotes the set of masked utterances; $\theta$, $\phi$, $\mathbf{W}$, and $\mathbf{b}$ are training parameters.

\subsubsection{Distributed Utterance Order Ranking (DUOR)}

Coherence is an essential aspect of conversation modeling. In a coherent discourse, utterances should respect specific orders of relations and logic. The ordering of utterances in a context determines the semantic of the conversation. Therefore, we hypothesize that learning to order a set of disordered utterances in such a way that maximizes the discourse coherence will have a critical impact in learning the representation of dialogue contexts. 

\begin{figure} [!tb]
		\centering 
		\includegraphics[width=3in]{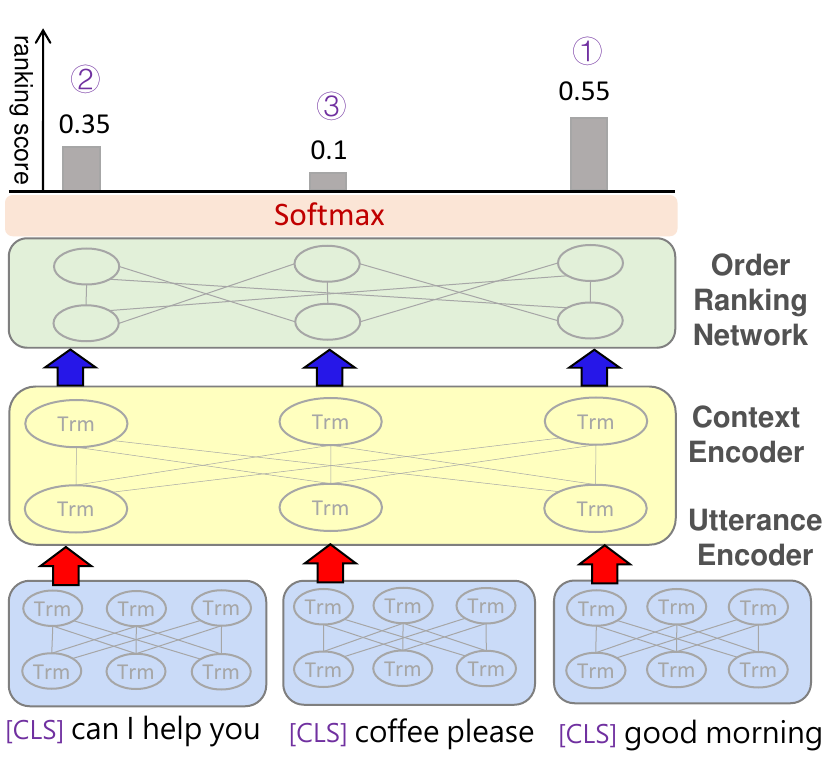} 
		\caption{Illustration of distributed utterance order ranking.}
		\label{fig:duor}
	\end{figure}

The goal of the utterance re-ordering task is to organize randomly shuffled utterances of a conversation into a coherent dialogue context~\citep{kumar2019ranktxnet}.
Formally, given a context $\mathcal{C}$ = $[u_{o_1}, u_{o_2},\ldots, u_{o_{|\mathcal{C}|}}]$ with order $o$ = $[o_1, o_2,\ldots , o_{|\mathcal{C}|}]$, we want an ordered context $\mathcal{C}^*$ = $[u_{o_1^*}, u_{o_2^*}, ..., u_{o_{|\mathcal{C}|}^*}]$ where $o^*$ =$[o_1^*, o_2^*,\ldots, o_{|\mathcal{C}|}^*]$ is the most coherent permutation of utterances. For example, the correct order for the utterances in Figure~\ref{fig:duor} is $o^*$ = $[3, 1, 2]$.

Previous work~\citep{sun2019ernie} models the sentence order prediction as a task of permutation index classification, namely, categorizing the index of the original permutation among $k$ classes where $k$ = $\sum_{n=1}^{|u|} n!$. However, the explosive permutation space for sets of utterances is often huge and largely overlapping (e.g., a slight modification to a permutation could lead to a far different class), 
leading to poor performance in current machine learning-based classifiers~\citep{luo2019gsoftmax,agrawal2013extremeclassification}.
	
In this paper, we design a distributed order ranking network (DORN) on top of the context encoder. Instead of categorizing permutation indices conditioned on the overall encoding of the context, DORN predicts the order index of each utterance in a distributed manner. As Figure~\ref{fig:duor} shows, DORN takes as input the hidden status of the shuffled utterances from the context encoder and produces a score for each individual utterance. These scores are then used for re-ordering these utterances (i.e., sorting these scores provides the correct ordering in the context). Inspired by the self-attention mechanism, the order prediction network computes the pairwise inner products between hidden states and then calculates the score~$s_i$ for each utterance~$u_i$ by averaging all its inner products to other utterances:
\begin{equation}
    s_i = \frac{1}{|\mathcal{C}|}\sum_{j=1}^{|\mathcal{C}|} \mathbf{W}\mathbf{h}_i^T\mathbf{h}_j
\end{equation}
where $\mathbf{W}$ denotes the parameters of DORN. 

In the training phase, we employ the \emph{learning to rank}~\citep{cao2007listnet} framework. Specifically, we view the predicted scores as the extent to which each utterance is ranked in the first place in a context. Given these scores, we estimate the ``rank-1''probability for each utterance with softmax: 
    \begin{equation}
        \hat{P}(u_i) = \frac{exp(s_i)}{\sum_{j=1}^{|\mathcal{C}|}exp(s_j)}
    \end{equation}
Accordingly, we assign a gold target value to each utterance to indicate the ground truth order. 
Then, the ``rank-1'' probability estimated by the gold target values is given as:
    \begin{equation}
        P(u_i) = \frac{exp(y_i)}{\sum_{j=1}^{|\mathcal{C}|}exp(y_j)}
    \end{equation}
    where $y_i\in[0,1]$ is the gold score for utterance~$u_i$. We set $y_i$ = $i/|\mathcal{C}|$ in our implementation.
Our goal is to minimize the KL divergence between the two distributions:
    \begin{equation}
      \begin{split}
        \mathcal{L}_\mathrm{duor}(\theta,\phi,\mathbf{W}| & P, \mathcal{C}) = \mathrm{KL}(\hat{P}(u)||P(u)) \\
         =  &\sum_{k=1}^{|\mathcal{C}|}\hat{P}(u_k)\mathrm{log}(\frac{\hat{P}(u_k)}{P(u_k)})
      \end{split}
    \end{equation}
where $\theta$, $\phi$ and $\mathbf{W}$ denote training parameters.

Overall, our objective is defined as the weighted sum of each loss function:
\begin{equation}
    \mathcal{L}_{total} = \mathcal{L}_{dec} + \lambda _{0} \times \mathcal{L}_{mur} + \lambda _{1} \times \mathcal{L}_{duor},
\end{equation}
where $\lambda_0$ and $\lambda_1$ denote the coefficients of each loss. We empirically set both $\lambda _{0}$ and $\lambda _{1}$ to 1 in our experiments.

\section{Experimental Setup}
\subsection{Dataset}
We evaluate DialogBERT using the following datasets (statistics shown in Table~\ref{table:dataset_overview}):
\begin{table}
\centering
\small
\begin{tabular}{c|ccc}
\toprule
\textbf{Dataset} & \bf Weibo & \bf MultiWOZ & \bf DailyDialog  \\ 
\midrule
conversations & 5,040,000 & 8,438 & 13,118 \\
total turns  & 20,521,891 & 113,556 & 102,979 \\
train samples & 15,481,891 & 106,794 & 76,052\\
valid samples & 89,994 & 12,902 & 7,069 \\
test samples  & 84,052 & 12,914  & 6,740 \\
\bottomrule
\end{tabular}
\caption{ Overview of the datasets}
\label{table:dataset_overview}
\end{table}

\noindent \textbf{Weibo} is a large-scale multi-turn conversation benchmark, introduced in the NLPCC2018 task5\footnote{http://tcci.ccf.org.cn/conference/2018/dldoc/taskgline05.pdf}.
The dataset originates from the Sina Weibo (microblog), a well-known Chinese Twitter-equivalent service, a popular data source for training conversation models~\citep{zeng2019you}. 

\noindent \textbf{MultiWOZ}\footnote{https://github.com/budzianowski/multiwoz} is a dataset that contains human-to-human written conversations spanning over multiple domains and topics~\citep{budzianowski2018multiwoz}. 
The dataset contains 10k dialogues. 
Although designed for task-oriented conversations, its daily-chatting style makes it suitable for evaluating open-domain conversations~\cite{mehri2019pretraining}.

\noindent \textbf{DailyDialog}\footnote{http://yanran.li/dailydialog} is a popular dataset for evaluating open-domain dialogue generation models~\citep{shen2018cvae_co,gu2019dialogwae}. The dataset contains multi-turn daily English dialogues suitable for English learners. 
Compared to Weibo and MultiWOZ, DailyDialog includes more chit-chat style utterances.

\begin{table*}[!htb]
\centering
\begin{tabular}{l|ccc|ccc|ccc}
\toprule
\multirow{2}{*}{\textbf{Model}} & \multicolumn{3}{c|}{Weibo} &  \multicolumn{3}{c|}{DailyDialog}&  \multicolumn{3}{c}{MultiWOZ} \\
\cline{2-10}  & \bf PPL & \bf BLEU & \bf NIST & \bf PPL & \bf BLEU & \bf NIST & \bf PPL & \bf BLEU & \bf NIST \\ 
\midrule
BART & 34.15 & 6.5 & 6.6 & 22.92 & 7.59 & 9.78  & 5.28 & 11.83 & 19.69 \\
DialoGPT & 87.14 & 7.24 & 8.28 & 46.42 & 13.82 & 14.08 & 9.44 & 18.47 & 34.15 \\
ContextPretrain-default  & 66.13 & 7.98 & 9.41 & 35.02 & 10.52 & 15.60 & 8.48 & 18.12 & 38.67 \\
ContextPretrain-transformer  & 36.92 & 8.10 & 9.17 & 22.31 &  10.14 & 13.55 & 5.52 & 17.95 & 37.16 \\ 
\midrule
NUG only& 31.73 & 8.25 & 9.45 & 22.16 & 14.41 & 22.33 & 5.06 & 19.50 & 49.83 \\
NUG+MUR & 28.08  & 8.49 & 10.12 & 21.14  & 13.81 & 21.50 & 4.96 & 19.89 & 52.74  \\
NUG+DUOR & \bf 27.29  & 8.46 & 10.04 & 20.82 & 14.42 & 22.65 & 5.00  & 19.87 & 53.29 \\
NUG+MUR+DUOR & 27.30  & \bf 8.54 & \bf 10.21 &\bf 20.49 &\bf 14.61 &\bf 23.34 &\bf 4.92 &\bf 19.91 &\bf 55.03\\
\bottomrule
\end{tabular}
\caption{Comparison between DialogBERT (including 4 variants in the bottom 4 rows) and baseline models on three datasets. We trained a base-size model in Weibo and small-size models in DailyDialog and MultiWOZ respectively. (NUG= Hierarchical BERT with NUG loss only, MUR=Masked Utterance Regression, DUOR=Distributed Utterance Order Ranking)}
\label{table:results:auto}
\end{table*}

\begin{table}[!tb]
\centering
\begin{tabular}{l@{}|@{}c@{}c@{}c@{}|@{}c@{}c@{}c}
\toprule
\multirow{2}{*}{\textbf{Size}} & \multicolumn{3}{c|}{MultiWOZ} & \multicolumn{3}{c}{DailyDialog} \\ \cline{2-7}
      & \bf \;PPL\; & \bf BLEU\; & \bf NIST\; & \bf \;PPL\; & \bf BLEU\; & \bf NIST \\ 
\hline
$\mathrm{L=12}$\; & \;4.90\; & 19.64 & 54.84\; & \;20.78 & 14.46 & 22.70\\
$\mathrm{L=9}$ & 4.91 & 19.19 & 54.74\; & \;20.67 & 14.55 & 23.17\\
$\mathrm{L=6}$ & 4.92 & 19.91 & 55.03\; & \;20.49 & 14.61 & 23.34\\
$\mathrm{L=3}$ & 4.94 & 19.19 & 52.48\; & \;20.85 & 13.89 & 21.57\\
\bottomrule
\end{tabular}
\caption{Performance of context encoder with difference sizes.}
\label{table:results:ablation1}
\end{table}

\begin{table}[!tb]
\centering
\begin{tabular}{l@{}|@{}c@{}c@{}c@{}|@{}c@{}c@{}c}
\toprule
\multirow{2}{*}{\textbf{Model}} & \multicolumn{3}{c|}{MultiWOZ} & \multicolumn{3}{c}{DailyDialog} \\ \cline{2-7}
 & \bf \;PPL\; & \bf BLEU\; & \bf NIST\;  & \bf \;PPL\; & \bf BLEU\; & \bf NIST\\ 
\hline
MURetr & 4.98 & 19.69 & 51.25\; & \;20.05 & 13.98 & 21.20\\
\textbf{MURegr\;} & \textbf{4.96} & \textbf{19.89} & \textbf{52.74\;} & \textbf{\;21.14}  & \textbf{13.81} & \textbf{21.50}\\
\hline
UOC & 5.15  & 18.68  & 48.07\; & \;23.85 & 10.58 & 14.90 \\
\textbf{DUOR} & \textbf{5.00} & \textbf{19.89} & \textbf{53.29\;} & \textbf{\;20.82} & \textbf{14.42} & \textbf{22.65}\\
\bottomrule
\end{tabular}
\caption{
Efficacy of our objectives compared to conventional ones (MURetr = Masked Utterance Retrieval, MURegr = Masked Utterance Regression, UOC= Utterance Order Classification, DUOR=Distributed Utterance Order Ranking). Bold fonts represent our proposed objectives.}
\label{table:results:ablation2}
\end{table}

\begin{table*}[htb]
\centering
\begin{tabular}{l|ccc|ccc|ccc}
\toprule
\multirow{2}{*}{\textbf{Model}} & \multicolumn{3}{c|}{Weibo} &  \multicolumn{3}{c|}{DailyDialog}&  \multicolumn{3}{c}{MultiWOZ} \\
\cline{2-10}  & \bf Size & \bf Train & \bf Test & \bf Size & \bf Train & \bf Test & \bf Size & \bf Train & \bf Test \\ 
\midrule
BART & 139.4M & 58h & 11h & 24.2M & 2.8h &  0.9h   & 24.2M & 6.6h & 1.8h \\
DialoGPT & 102.1M & 27.5h  & 9.8h  & 12.7M & 2.6h & 0.85h  & 12.7M & 8.9h & 1.8h \\
ContextPretrain  & 20.5M & 14.6h  & 21h  & 23.3M & 16.8h & 1.4h & 23.3M & 17.9h & 3.2h \\
\midrule
DialogBERT& 337.6M & 27.9h & 23.3h & 40.2M &  9.0h & 1.3h  & 40.2M & 9.8h & 1.9h \\
\bottomrule
\end{tabular}
\caption{Comparison of model size (i.e., number of parameters), training and inference time between DialogBERT and baseline models in terms of three datasets (M: million, h: hours).}
\label{table:results:efficiency}
\end{table*}

\subsection{Implementation \& Reproducibility}
We implemented our approach on top of the Huggingface Transformer repository~\citep{Wolf2019HuggingFacesTS}. 
We trained two different sizes of the model to accommodate to the datasets. 
For the Weibo dataset, we trained a base-sized model in which both encoders (including the utterance encoder and the context encoder) and the decoder use the hyper-parameter settings of `\emph{bert-base-chinese}'~(L=12, H=768, A=12). Since the base-size model is vulnerable to over-fitting in small datasets, we also trained small-sized models on MultiWOZ and DailyDialog.
The small-size model reduces the `\emph{bert-base-uncased}' configuration to 6 transformer layers, has a hidden size of 256, and contains 2 attention heads (L=6, H=256, A=2).
We limit the number of utterances in each context to 7~\citep{adiwardana2020meena} and the utterance length to 30 words. 
All of the experiments use the default BERT tokenizer (e.g., \emph{bert-base-uncased} for English datasets).
All models were optimized with AdamW~\citep{loshchilov2018adamw} optimizer using an initial learning rate of 5$e^{-5}$. 
We used the adaptive learning-rate scheduler with 5,000 warm-up steps. 
We implemented all the models with the PyTorch library. 
Experiments took place on a machine with Ubuntu 16.04 and an NVidia Tesla P40 GPU. 

We compute the validation loss every 2,000 iterations and choose the best set of parameters. 
The test scores from the best checkpoint represent the final results. 
We repeat the experiments five times and report the average results.
We search for hyperparameters using NAVER Smart Machine Learning (NSML)~\cite{sung2017nsml,park2019visualhypertuner}.

During response generation, we perform top-1 sampling according to the probabilities estimated by Equation~\ref{eq:decoder}.

\subsection{Baseline Models}
We compare our approach with previous Transformer-based response generation methods. (i) BART~\citep{lewis2019bart} is a pre-trained Transformer encoder-decoder model for conditional text generation. 
BART can be seen as a generalized BERT (due to the bidirectional encoder) and GPT (with the left-to-right decoder). 
As such, BART is prevalent in conversational systems~\citep{adiwardana2020meena}.
(ii) DialoGPT~\citep{zhang2019dialogpt} is a pre-training language model for conversation generation based on GPT-2~\citep{radford2019language}.
(iii) ContextPretrain~\citep{mehri2019pretraining} refers to a set of pre-training methods for dialogue context representation learning. The paper proposes masked context retrieval and inconsistent utterance identification. To control the expressive power, we use the Transformer-based decoder in place of the RNN decoder. To ensure fairness, we used the same hyperparameter settings (e.g., the number of transformer layers and attention heads) for all baseline models.

\subsection{Evaluation Metrics}
Following~\cite{adiwardana2020meena}, we use
perplexity to measure the performance of response generation. 
Studies \citep{adiwardana2020meena} show that this widely-used metric correlates with human judgment scores significantly.
We further adopt two commonly used metrics in related works including BLEU~\citep{papineni2002bleu} and NIST~\citep{doddington2002nist}.
BLEU\footnote{https://www.nltk.org/\_modules/nltk/translate/bleu\_score.html} measures how many n-grams in a generated response overlap with those of the reference. 
We calculate BLEU-4 scores in our experiments. 
NIST~\citep{doddington2002nist} is a variant of BLEU that penalizes uninformative n-grams by assigning weights to n-grams according to their information gain\footnote{https://www.nltk.org/\_modules/nltk/translate/nist\_score.html}.

\section{Evaluation Results}

\subsection{Automatic Evaluation}
Table~\ref{table:results:auto} shows the performance of each method on the automatic metrics. 
To demonstrate the effect of the proposed training objectives, we present the results of DialogBERT by ablating different combinations of the objectives. 
Broadly, DialogBERT (pre-trained with all proposed training objectives) achieves the best performance on most automatic metrics, especially for the perplexity.
Compared to models with flat context encoding such as BART and DialoGPT, DialogBERT outperforms baseline models on all metrics with a large margin. 
Such improvements are consistent across all three datasets, affirming the superiority of the hierarchical Transformer architecture of DialogBERT.

The results show that both the proposed masked utterance regression and distributed utterance order prediction achieve substantial improvements over a simple hierarchical Transformer.
Furthermore, combining both objectives further enhances the performance. 
Interestingly, the improvement on the Weibo dataset is relatively more significant.
We conjecture that this is due to the richness of the data, allowing more room for fitting using the auxiliary objectives.

In all three datasets, DialoGPT performs considerably worse than the other methods regarding the perplexity. 
Presumably, this is due to the the GPT-2 language model's auto-regressive nature and the single-turn setting~\citep{zhang2019dialogpt}. 
We further discuss this issue in the case study.

To investigate the efficacy of the context encoder, we conduct an ablation study on the encoder size. 
We train DialogBRET with various numbers of Transformer layers in the context encoder. 
As shown in Table~\ref{table:results:ablation1}, the number of Transformer layers in the context encoder has a modest effect on the performance, which indicates the ease of extracting context representations. We conjecture that this may be because the discourse-level interaction is relatively easier to capture than the word-level semantics.

We also investigated the effects of two proposed training objectives, which we compare with previously-proposed objectives for paragraphs, namely the masked utterance retrieval~\citep{mehri2019pretraining} and the utterance order classification~\citep{sun2019ernie}. 
As Table~\ref{table:results:ablation2} shows, the proposed MURegr performs competitively with the \emph{masked utterance retrieval} objective, supporting the idea that the MURegr is a simple yet effective substitute for dialog context representation. 
One explanation is that MURegr offers more fine-grained representation learning than retrieval-based objectives (negative utterance classification) thanks to regression (exact vector matching).
Thus, for missing utterance reconstruction, the goal of exact matching (regression) becomes more impactful than the goal of classification.
Meanwhile, the proposed DUOR objective performs significantly better than the existing \emph{utterance order classification} objective.

We proceed to analyze the computational efficiency. 
Table~\ref{table:results:efficiency} shows the model size, training and test time for each model.
Although DialogBERT has a larger model size due to an extra ``context encoder'', it does not incur significant overhead since the length (the number of utterances in each dialogue) of the context encoder input is usually short.

\begin{table*} [!htb]
\centering
\begin{tabular}{l|ccc|ccc|ccc}
\toprule 
\multirow{2}{*}{\textbf{Comparison}} &
    \multicolumn{3}{c|}{{Coherence}} & \multicolumn{3}{c|}{{Informativeness}} & \multicolumn{3}{c}{{Human-likeness}} \\
\cline{2-10}
 & \textbf{Win} & \textbf{Tie} & \textbf{Loss} &
    \textbf{Win} & \textbf{Tie} & \textbf{Loss} &
    \textbf{Win} & \textbf{Tie} & \textbf{Loss} \\
\midrule
Ours \textit{vs.} BART & 41\% & 34\% & 25\% & 38\% & 43\% & 19\% & 39\% & 45\% & 15\% \\
Ours \textit{vs.} DialoGPT & 39\% & 45\% & 16\% & 40\% & 48\% & 12\%  & 49\% & 43\% & 8\% \\
Ours \textit{vs.} ContextPretrain & 44\% & 34\% & 22\% & 40\% & 42\% & 18\% & 39\% & 42\% & 19\% \\
\bottomrule
\end{tabular}
\caption{Human evaluation on the test set of DailyDialog. The NUG+MUR+DUOP in Table~\ref{table:results:auto} is selected for comparison. All results have statistical significance of $p<0.001$.}
\label{table:results:human}
\end{table*}

\begin{table*} [!htb]
\small
\centering
\begin{tabular}{rl}
\toprule
\multirow{4}{*}{\textbf{Context}} & A: I'm looking for a college in town.\\ 
                             & B: Is there a particular area you're looking for a college in? christ's college is \\
                             & ~~~~~in the centre area, it is located on saint andrew.\\
                             & A: Could you give me the address for it?   \\
                             \hline\hline
  \bf BART: & Sure. \\ 
  \bf DialoGPT: & Yes will, give thecode thecode. post is21,21 and entrance is. address \\
                & ~ saint'street pose cbbu and admission free \\
  \bf ContextPretrain:  & Yes, it is moderately priced and do you want free parking and wifi? \\ 
  \bf DialogBERT: & \textbf{Their address is 98 king street and their post code is cb11ln. Can I help} \\
           & \textbf{~you with anything else?} \\
  \bf Human: & The address is saint andrew's street and the postcode is cb23bu. Is there \\
              & anything else I can help you with? \\
\bottomrule
\end{tabular}

\begin{tabular}{rl}
\toprule
\multirow{7}{*}{\textbf{Context}} & A: Then how much is that by airmail? \\
       & B: Let me see. it's 57 dollars, 20 cents, including tax.\\
       & A: I guess I have to send it by airmail.\\
       & B: Do you want to ensure the contents, miss?\\
       & A: Yes, please. \\ 
\hline\hline
\textbf{BART}: & I would your e mail.\\ 
\textbf{DialoGPT}: & Thank. only one much.\\ 
\textbf{ContextPretrain}:  & Do you need any discounts? \\ 
\textbf{DialogBERT}: & \textbf{Would you please write the number and tell me the best?} \\
\textbf{Human}: & Please fill out this form, also please write the value of the items in this space.\\
\bottomrule
\end{tabular}
\caption{Sample conversations from multiple models with human reference}
\label{table:case}
\end{table*}

\subsection{Human Evaluation}
We conduct a human evaluation using the Amazon Mechanical Turk platform. 
We used the DailyDialog as the evaluation corpus since its properties (daily chit-chats) make it easier for annotators to judge.
We randomly sampled 200 dialogues to be judged by the workers.
For each, we presented (i) the entire dialogue context, (ii) the response generated by our model, and (iii) the response generated by the competing models without disclosing the source models.
For each pair of sample responses, we asked three different annotators to blindly evaluate the quality regarding the three criteria (coherence, informativeness and human-likeness) and express their preference using a 3-point Likert scale: ``win'' (ours is better), ``loss'' (the other is better) and ``tie'' (equally good or bad). 
We screened the workers by qualifications, and we also filtered out low-quality answers using spam detection methods.
Table~\ref{table:results:human} shows the overall distribution of the answers.
Overall, the workers preferred our model over other baselines regarding the relevance to the dialogue context, informativeness, and human-likeness.


\subsection{Case Study}
  \label{ss:results:case}

We provide two generated dialogues in Table~\ref{table:case} from the MultiWOZ and DailyDialog datasets.
The two samples illustrate that DialogBERT generates more coherent responses than the baseline models, consistent with our automatic and human evaluation results.
Interestingly, DialogBERT exhibits \emph{meticulousness} that makes its response more human-like.
For example, the first sample shows that DialogBERT produces very detailed responses such as the ``\emph{98 king street}'' and the ``\emph{post code cb1ln}'' instead of vague ones observed in samples from other models.
This observation is in agreement with our human study on informativeness and human-likeness. 

Furthermore, in both examples, DialoGPT produces less relevant results than other methods.
Again, this is related to the single-turn setting~\citep{zhang2019dialogpt}. 
Specifically, DialoGPT treats response generation as a pure auto-regressive language model. 
The entire context is taken as input to generate subsequent words step by step, making it difficult to handle multiple turns, especially in relatively small conversation datasets.


\section{Conclusion}
In this paper, we proposed a neural response generation model named DialogBERT. 
Instead of encoding the dialogue context as a linear sequence of tokens, DialogBERT employs a hierarchical Transformer encoder architecture. 
As a natural extension of the original BERT training, we proposed two training objectives: masked utterance regression and distributed utterance re-ordering.
We showed that the proposed objectives enable the conversation model to capture multi-level (discourse-level and utterance-level) coherences.
Additionally, we showed that DialogBERT notably outperforms baseline models on the response generation tasks.

\section*{Acknowledgments}
The authors would thank Prof. Kyunghyun Cho at New York University for his valuable comments on this project. This work was done when the first author was visiting NAVER AI Lab.

\bibliography{references}

\begin{thebibliography}{38}
\providecommand{\natexlab}[1]{#1}
\providecommand{\url}[1]{\texttt{#1}}
\providecommand{\urlprefix}{URL }
\expandafter\ifx\csname urlstyle\endcsname\relax
  \providecommand{\doi}[1]{doi:\discretionary{}{}{}#1}\else
  \providecommand{\doi}{doi:\discretionary{}{}{}\begingroup
  \urlstyle{rm}\Url}\fi

\bibitem[{Adiwardana et~al.(2020)Adiwardana, Luong, So, Hall, Fiedel,
  Thoppilan, Yang, Kulshreshtha, Nemade, Lu et~al.}]{adiwardana2020meena}
Adiwardana, D.; Luong, M.-T.; So, D.~R.; Hall, J.; Fiedel, N.; Thoppilan, R.;
  Yang, Z.; Kulshreshtha, A.; Nemade, G.; Lu, Y.; et~al. 2020.
\newblock Towards a human-like open-domain chatbot.
\newblock \emph{arXiv preprint arXiv:2001.09977} .

\bibitem[{Agrawal et~al.(2013)Agrawal, Gupta, Prabhu, and
  Varma}]{agrawal2013extremeclassification}
Agrawal, R.; Gupta, A.; Prabhu, Y.; and Varma, M. 2013.
\newblock Multi-label learning with millions of labels: Recommending advertiser
  bid phrases for web pages.
\newblock In \emph{Proceedings of the 22nd international conference on World
  Wide Web}, 13--24.

\bibitem[{Budzianowski and Vuli{\'c}(2019)}]{budzianowski2019hello}
Budzianowski, P.; and Vuli{\'c}, I. 2019.
\newblock Hello, It’s GPT-2-How Can I Help You? Towards the Use of Pretrained
  Language Models for Task-Oriented Dialogue Systems.
\newblock In \emph{Proceedings of the 3rd Workshop on Neural Generation and
  Translation}, 15--22.

\bibitem[{Budzianowski et~al.(2018)Budzianowski, Wen, Tseng, Casanueva, Ultes,
  Ramadan, and Ga{\v{s}}i{\'c}}]{budzianowski2018multiwoz}
Budzianowski, P.; Wen, T.-H.; Tseng, B.-H.; Casanueva, I.; Ultes, S.; Ramadan,
  O.; and Ga{\v{s}}i{\'c}, M. 2018.
\newblock Multiwoz-a large-scale multi-domain wizard-of-oz dataset for
  task-oriented dialogue modelling.
\newblock \emph{arXiv preprint arXiv:1810.00278} .

\bibitem[{Cao et~al.(2007)Cao, Qin, Liu, Tsai, and Li}]{cao2007listnet}
Cao, Z.; Qin, T.; Liu, T.-Y.; Tsai, M.-F.; and Li, H. 2007.
\newblock Learning to rank: from pairwise approach to listwise approach.
\newblock In \emph{Proceedings of the 24th international conference on Machine
  learning}, 129--136.

\bibitem[{Cho et~al.(2014)Cho, Van~Merri{\"e}nboer, Gulcehre, Bahdanau,
  Bougares, Schwenk, and Bengio}]{cho2014seq2seq}
Cho, K.; Van~Merri{\"e}nboer, B.; Gulcehre, C.; Bahdanau, D.; Bougares, F.;
  Schwenk, H.; and Bengio, Y. 2014.
\newblock Learning phrase representations using RNN encoder-decoder for
  statistical machine translation.
\newblock \emph{arXiv preprint arXiv:1406.1078} .

\bibitem[{Clark et~al.(2020)Clark, Luong, Le, and Manning}]{clark2020electra}
Clark, K.; Luong, M.-T.; Le, Q.~V.; and Manning, C.~D. 2020.
\newblock ELECTRA: Pre-training Text Encoders as Discriminators Rather Than
  Generators.
\newblock In \emph{International Conference on Learning Representations}.

\bibitem[{Devlin et~al.(2018)Devlin, Chang, Lee, and
  Toutanova}]{devlin2018bert}
Devlin, J.; Chang, M.-W.; Lee, K.; and Toutanova, K. 2018.
\newblock Bert: Pre-training of deep bidirectional transformers for language
  understanding.
\newblock \emph{arXiv preprint arXiv:1810.04805} .

\bibitem[{Doddington(2002)}]{doddington2002nist}
Doddington, G. 2002.
\newblock Automatic evaluation of machine translation quality using n-gram
  co-occurrence statistics.
\newblock In \emph{Proceedings of the second international conference on Human
  Language Technology Research}, 138--145.

\bibitem[{Gu et~al.(2019)Gu, Cho, Ha, and Kim}]{gu2019dialogwae}
Gu, X.; Cho, K.; Ha, J.-W.; and Kim, S. 2019.
\newblock DialogWAE: Multimodal Response Generation with Conditional
  Wasserstein Auto-Encoder.
\newblock In \emph{Proceedings of the 7th International Conference on Learning
  Representations (ICLR 2019)}.

\bibitem[{Henderson et~al.(2019)Henderson, Casanueva, Mrk{\v{s}}i{\'c}, Su,
  Vuli{\'c} et~al.}]{henderson2019convert}
Henderson, M.; Casanueva, I.; Mrk{\v{s}}i{\'c}, N.; Su, P.-H.; Vuli{\'c}, I.;
  et~al. 2019.
\newblock ConveRT: Efficient and Accurate Conversational Representations from
  Transformers.
\newblock \emph{arXiv preprint arXiv:1911.03688} .

\bibitem[{Kumar et~al.(2019)Kumar, Brahma, Karnick, and
  Rai}]{kumar2019ranktxnet}
Kumar, P.; Brahma, D.; Karnick, H.; and Rai, P. 2019.
\newblock Deep Attentive Ranking Networks for Learning to Order Sentences.
\newblock \emph{arXiv preprint arXiv:2001.00056} .

\bibitem[{Lewis et~al.(2019)Lewis, Liu, Goyal, Ghazvininejad, Mohamed, Levy,
  Stoyanov, and Zettlemoyer}]{lewis2019bart}
Lewis, M.; Liu, Y.; Goyal, N.; Ghazvininejad, M.; Mohamed, A.; Levy, O.;
  Stoyanov, V.; and Zettlemoyer, L. 2019.
\newblock Bart: Denoising sequence-to-sequence pre-training for natural
  language generation, translation, and comprehension.
\newblock \emph{arXiv preprint arXiv:1910.13461} .

\bibitem[{Loshchilov and Hutter(2018)}]{loshchilov2018adamw}
Loshchilov, I.; and Hutter, F. 2018.
\newblock Decoupled Weight Decay Regularization.
\newblock In \emph{International Conference on Learning Representations}.

\bibitem[{Luo et~al.(2019)Luo, Wong, Kankanhalli, and Zhao}]{luo2019gsoftmax}
Luo, Y.; Wong, Y.; Kankanhalli, M.; and Zhao, Q. 2019.
\newblock G-Softmax: Improving Intraclass Compactness and Interclass
  Separability of Features.
\newblock \emph{IEEE transactions on neural networks and learning systems} .

\bibitem[{Mehri et~al.(2019)Mehri, Razumovskaia, Zhao, and
  Eskenazi}]{mehri2019pretraining}
Mehri, S.; Razumovskaia, E.; Zhao, T.; and Eskenazi, M. 2019.
\newblock Pretraining Methods for Dialog Context Representation Learning.
\newblock In \emph{Proceedings of the 57th Annual Meeting of the Association
  for Computational Linguistics}, 3836--3845.

\bibitem[{Melamud, Bornea, and Barker(2019)}]{melamud2019combining}
Melamud, O.; Bornea, M.; and Barker, K. 2019.
\newblock Combining Unsupervised Pre-training and Annotator Rationales to
  Improve Low-shot Text Classification.
\newblock In \emph{Proceedings of the 2019 Conference on Empirical Methods in
  Natural Language Processing and the 9th International Joint Conference on
  Natural Language Processing (EMNLP-IJCNLP)}, 3875--3884.

\bibitem[{Ortega and Vu(2017)}]{ortega2017neural}
Ortega, D.; and Vu, N.~T. 2017.
\newblock Neural-based context representation learning for dialog act
  classification.
\newblock \emph{arXiv preprint arXiv:1708.02561} .

\bibitem[{Papineni et~al.(2002)Papineni, Roukos, Ward, and
  Zhu}]{papineni2002bleu}
Papineni, K.; Roukos, S.; Ward, T.; and Zhu, W.-J. 2002.
\newblock BLEU: a method for automatic evaluation of machine translation.
\newblock In \emph{Proceedings of the 40th annual meeting on association for
  computational linguistics}, 311--318. Association for Computational
  Linguistics.

\bibitem[{Park et~al.(2019)Park, Kim, Kim, Kim, Choo, Ha, and
  Sung}]{park2019visualhypertuner}
Park, H.; Kim, J.; Kim, M.; Kim, J.-H.; Choo, J.; Ha, J.-W.; and Sung, N. 2019.
\newblock VisualHyperTuner: Visual analytics for user-driven hyperparameter
  tuning of deep neural networks.
\newblock In \emph{Demo at SysML Conference}.

\bibitem[{Pennington, Socher, and Manning(2014)}]{pennington2014glove}
Pennington, J.; Socher, R.; and Manning, C.~D. 2014.
\newblock Glove: Global vectors for word representation.
\newblock In \emph{Proceedings of the 2014 conference on empirical methods in
  natural language processing (EMNLP)}, 1532--1543.

\bibitem[{Peters et~al.(2018)Peters, Neumann, Iyyer, Gardner, Clark, Lee, and
  Zettlemoyer}]{peters2018deep}
Peters, M.; Neumann, M.; Iyyer, M.; Gardner, M.; Clark, C.; Lee, K.; and
  Zettlemoyer, L. 2018.
\newblock Deep Contextualized Word Representations.
\newblock In \emph{Proceedings of the 2018 Conference of the North American
  Chapter of the Association for Computational Linguistics: Human Language
  Technologies, Volume 1 (Long Papers)}, 2227--2237.

\bibitem[{Pragst et~al.(2018)Pragst, Rach, Minker, and
  Ultes}]{pragst2018vector}
Pragst, L.; Rach, N.; Minker, W.; and Ultes, S. 2018.
\newblock On the vector representation of utterances in dialogue context.
\newblock In \emph{Proceedings of the Eleventh International Conference on
  Language Resources and Evaluation (LREC 2018)}.

\bibitem[{Radford et~al.(2019)Radford, Wu, Child, Luan, Amodei, and
  Sutskever}]{radford2019language}
Radford, A.; Wu, J.; Child, R.; Luan, D.; Amodei, D.; and Sutskever, I. 2019.
\newblock Language Models are Unsupervised Multitask Learners.
\newblock \emph{Open AI Blog} 1: 9.

\bibitem[{Roller et~al.(2020)Roller, Dinan, Goyal, Ju, Williamson, Liu, Xu,
  Ott, Shuster, Smith et~al.}]{roller2020recipes}
Roller, S.; Dinan, E.; Goyal, N.; Ju, D.; Williamson, M.; Liu, Y.; Xu, J.; Ott,
  M.; Shuster, K.; Smith, E.~M.; et~al. 2020.
\newblock Recipes for building an open-domain chatbot.
\newblock \emph{arXiv preprint arXiv:2004.13637} .

\bibitem[{Rothe, Narayan, and Severyn(2019)}]{rothe2019bert2bert}
Rothe, S.; Narayan, S.; and Severyn, A. 2019.
\newblock Leveraging pre-trained checkpoints for sequence generation tasks.
\newblock \emph{arXiv preprint arXiv:1907.12461} .

\bibitem[{Shen et~al.(2018)Shen, Su, Niu, and Demberg}]{shen2018cvae_co}
Shen, X.; Su, H.; Niu, S.; and Demberg, V. 2018.
\newblock Improving variational encoder-decoders in dialogue generation.
\newblock In \emph{Thirty-Second AAAI Conference on Artificial Intelligence}.

\bibitem[{So, Liang, and Le(2019)}]{so2019evol}
So, D.~R.; Liang, C.; and Le, Q.~V. 2019.
\newblock The evolved transformer.
\newblock \emph{arXiv preprint arXiv:1901.11117} .

\bibitem[{Sun et~al.(2019)Sun, Wang, Li, Feng, Tian, Wu, and
  Wang}]{sun2019ernie}
Sun, Y.; Wang, S.; Li, Y.; Feng, S.; Tian, H.; Wu, H.; and Wang, H. 2019.
\newblock Ernie 2.0: A continual pre-training framework for language
  understanding.
\newblock \emph{arXiv preprint arXiv:1907.12412} .

\bibitem[{Sung et~al.(2017)Sung, Kim, Jo, Yang, Kim, Lausen, Kim, Lee, Kwak, Ha
  et~al.}]{sung2017nsml}
Sung, N.; Kim, M.; Jo, H.; Yang, Y.; Kim, J.; Lausen, L.; Kim, Y.; Lee, G.;
  Kwak, D.; Ha, J.-W.; et~al. 2017.
\newblock Nsml: A machine learning platform that enables you to focus on your
  models.
\newblock \emph{arXiv preprint arXiv:1712.05902} .

\bibitem[{Vaswani et~al.(2017)Vaswani, Shazeer, Parmar, Uszkoreit, Jones,
  Gomez, Kaiser, and Polosukhin}]{vaswani2017attention}
Vaswani, A.; Shazeer, N.; Parmar, N.; Uszkoreit, J.; Jones, L.; Gomez, A.~N.;
  Kaiser, {\L}.; and Polosukhin, I. 2017.
\newblock Attention is all you need.
\newblock In \emph{Advances in neural information processing systems},
  5998--6008.

\bibitem[{Wolf et~al.(2019{\natexlab{a}})Wolf, Debut, Sanh, Chaumond, Delangue,
  Moi, Cistac, Rault, Louf, Funtowicz, and Brew}]{Wolf2019HuggingFacesTS}
Wolf, T.; Debut, L.; Sanh, V.; Chaumond, J.; Delangue, C.; Moi, A.; Cistac, P.;
  Rault, T.; Louf, R.; Funtowicz, M.; and Brew, J. 2019{\natexlab{a}}.
\newblock HuggingFace's Transformers: State-of-the-art Natural Language
  Processing.
\newblock \emph{ArXiv} abs/1910.03771.

\bibitem[{Wolf et~al.(2019{\natexlab{b}})Wolf, Sanh, Chaumond, and
  Delangue}]{wolf2019transfertransfo}
Wolf, T.; Sanh, V.; Chaumond, J.; and Delangue, C. 2019{\natexlab{b}}.
\newblock Transfertransfo: A transfer learning approach for neural network
  based conversational agents.
\newblock \emph{arXiv preprint arXiv:1901.08149} .

\bibitem[{Yang et~al.(2019)Yang, Dai, Yang, Carbonell, Salakhutdinov, and
  Le}]{yang2019xlnet}
Yang, Z.; Dai, Z.; Yang, Y.; Carbonell, J.; Salakhutdinov, R.~R.; and Le, Q.~V.
  2019.
\newblock Xlnet: Generalized autoregressive pretraining for language
  understanding.
\newblock In \emph{Advances in neural information processing systems},
  5754--5764.

\bibitem[{Zeng et~al.(2019)Zeng, Li, He, Gao, Lyu, and King}]{zeng2019you}
Zeng, J.; Li, J.; He, Y.; Gao, C.; Lyu, M.~R.; and King, I. 2019.
\newblock What you say and how you say it: Joint modeling of topics and
  discourse in microblog conversations.
\newblock \emph{Transactions of the Association for Computational Linguistics}
  7: 267--281.

\bibitem[{Zhang, Wei, and Zhou(2019)}]{zhang2019hibert}
Zhang, X.; Wei, F.; and Zhou, M. 2019.
\newblock HIBERT: Document level pre-training of hierarchical bidirectional
  transformers for document summarization.
\newblock \emph{arXiv preprint arXiv:1905.06566} .

\bibitem[{Zhang et~al.(2019)Zhang, Sun, Galley, Chen, Brockett, Gao, Gao, Liu,
  and Dolan}]{zhang2019dialogpt}
Zhang, Y.; Sun, S.; Galley, M.; Chen, Y.-C.; Brockett, C.; Gao, X.; Gao, J.;
  Liu, J.; and Dolan, B. 2019.
\newblock DialoGPT: Large-Scale Generative Pre-training for Conversational
  Response Generation.
\newblock \emph{arXiv preprint arXiv:1911.00536} .

\bibitem[{Zheng et~al.(2020)Zheng, Zhang, Huang, and Mao}]{zheng2020pre}
Zheng, Y.; Zhang, R.; Huang, M.; and Mao, X. 2020.
\newblock A Pre-Training Based Personalized Dialogue Generation Model with
  Persona-Sparse Data.
\newblock In \emph{AAAI}, 9693--9700.

\end{thebibliography}
\bibliographystyle{aaai21}



\end{document}